\documentclass[
]{ceurart}

\usepackage{subcaption}
\usepackage{amsmath}

\begin{document}



\title{Does spatio-temporal information benefit the video summarization task?}


\copyrightyear{}
\copyrightclause{}

\conference{}
\author[1]{Aashutosh Ganesh}[%
email=Aashutosh.Ganesh@maastrichtuniversity.nl
]
\cormark[1]
\author[1]{Mirela Popa}[%
email=m.popa@maastrichtuniversity.nl
]
\author[2]{Daan Odijk}[%
email=Daan.Odjik@rtl.nl
]
\author[1]{Nava Tintarev}[%
email=n.tintarev@maastrichtuniversity.nl
]
\address[1]{Department of Advanced Computing Sciences, Maastricht University, the Netherlands}
\address[2]{RTL, Hilversum, the Netherlands}

\cortext[1]{Corresponding author.}
\begin{abstract}
An important aspect of summarizing videos is understanding the temporal context behind each part of the video to grasp what is and is not important.
Video summarization models have in recent years modeled spatio-temporal relationships to represent this information. These models achieved state-of-the-art correlation scores on important benchmark datasets. However, what has not been reviewed is whether spatio-temporal relationships are even required to achieve state-of-the-art results.
Previous work in activity recognition has found biases, by prioritizing static cues such as scenes or objects, over motion information. In this paper we inquire if similar spurious relationships might influence the task of video summarization. To do so, we analyse the role that temporal information plays on existing benchmark datasets. We first estimate a baseline with temporally invariant models to see how well such models rank on benchmark datasets (TVSum and SumMe). We then disrupt the temporal order of the videos to investigate the impact it has on existing state-of-the-art models. One of our findings is that the temporally invariant models achieve competitive correlation scores that are close to the human baselines on the TVSum dataset. We also demonstrate that existing models are not affected by temporal perturbations. Furthermore, with certain disruption strategies that shuffle fixed time segments, we can actually
improve their correlation scores. With these results, we find that spatio-temporal relationship play a minor role and we raise the question whether these benchmarks adequately model the task of video summarization.
Code available at: https://github.com/AashGan/TemporalPerturbSum
\end{abstract}

\maketitle

\section{Introduction}

A striking amount of short-form video content is created, hosted, and consumed within the online media landscape. Several platforms such as Tiktok, Youtube and Instagram promote these short snappy videos as they immediately capture the users' interest. These videos are often created by cutting a longer video into its best parts. However, the process of editing them into these bite-sized pieces is still time-intensive with significant potential for automation. One way to broach automatic editing is by using video summarization algorithms \cite{otani2022video}. %
\par
What makes a good video summary is however largely subjective and is dependent on the underlying media. Regardless of this subjectivity, a pivotal aspect in understanding what is pertinent for a video summary is the \textit{temporal context} behind different parts of a video. The context in this case is the relationship one part of a video shares with other parts, as the preceding or succeeding frames or shots inform us as to what may be relevant. The information gained by learning the temporal context gives us vital cues regarding what should be included in a summary. Therefore, we may expect automatic video summarization algorithms by design to discover and exploit such relationships.
\par 
Based on this assumption, modern approaches to video summarization utilise Deep Neural Networks, which employ spatio-temporal relationships within the data \cite{zhang2016video,memory-networks,pglsum,rochan2018video} to understand the temporal context within the videos. These approaches often estimate a frame-wise importance score, which indicates how likely is that a frame should be included in a summary. To evaluate the accuracy of their predictions, they measure how well these scores correlate with gold standard human labels provided by two important benchmark datasets: TVSum \cite{tvsum} and SumMe \cite{summe}. These datasets despite their small size, are a cornerstone within video summarization research due to their diverse set of videos and the inclusion of multiple human annotations per video to capture the subjectivity of video summarization.
\par
An important question that can be raised regarding these benchmark datasets is the role spatio-temporal relationship play within them. Despite the state-of-the-art Kendall rank correlation coefficient scores achieved by these spatio-temporal models, it has been observed by Otani et al \cite{otani2022video} that a simple multi-ranking model which does not account for temporal relationships by design, also achieved competitive scores. This casts a doubt regarding the trustworthiness of the aforementioned models to capture temporal context. 
\par
A similar question has also been posed within activity recognition research \cite{li2019repair,Li2018RESOUNDTA}. Some of the relevant works showcase how activity recognition models could effectively ignore the temporal component of the video, while still accurately predicting the activity. In some cases, this could be due to a ``representation bias" within the dataset. This bias, as illustrated by Li et al. \cite{Li2018RESOUNDTA}, is where a dataset may favour a certain data representation, influencing the model to learn spurious relationships. For instance, videos could be collected for an activity such as playing football, may have associated videos from a specific field. This type of intrinsic correlation might be captured by the DNN model, which recognizes correctly the activity class by detecting the field instead of the football related motion. This may give us the false impression that an activity recognition model should simply recognize the salient objects or scenes. This also poses a significant challenge to the trustworthiness of such systems, since it may use unexpected and spurious relationships for predicting a certain activity class.
\par
An investigation of the role temporal dependencies play within popular video summarization benchmark datasets is important in order to evaluate the trustworthiness of these methods. If a temporally invariant model achieves a performance that is on par with its temporally dependent counterparts, then this may highlight an issue within the benchmarks used to compare them. To analyse this hypothesis, experiments which manipulate the temporal order of a video will indicate whether these dependencies are truly needed. Furthermore, interesting insights could be formulated with respect to the amount of data needed to capture various temporal relationships. These observations highlight the growing need for scrutiny over benchmark datasets, by critically assessing whether the underlying data distribution is adequate to model the intended types of relationships, leading to increasing trust in AI systems.

\par

Therefore in this work we investigate the role spatio-temporal dependencies play on the video summarization benchmark datasets. We achieve this goal 
by disrupting the temporal order of the videos. We first establish a baseline performance using temporally invariant models on TVSum and SumMe datasets. On this baseline, we investigate the effect that temporal disruptions have on models which utilise spatio-temporal relationships. We take inspiration from time series data augmentation \cite{um2017data} to design different temporal disruptions introduced at different timescales, including low-level, intermediate and global levels, represented by frames and shots. Through these experiments, we investigate the role played by the temporal context in video summarization on two important benchmark datasets. 

\par
To summarise, in this paper we make the following contributions:
\begin{itemize}
    \item We demonstrate that models which do not utilise temporal context can still achieve close to the state-of-the-art correlation scores on the TVSum and SumMe datasets. 
    \item We highlight that the introduction of temporal disruptions had a limited effect on the performance of video summarization models on the two considered benchmark datasets.  Moreover, we also prove that the analyzed temporal disruptions in some cases even improved the models' performance, underscoring the role played by spatio-temporal relationships.
    \item Finally, we trace back the aforementioned results to the design of the datasets, the architecture and the evaluation stage. Our results indicate the need to address the identified limitations, when designing future benchmarks in video summarization, by properly evaluating the expected contributions at various temporal levels. 
    \end{itemize}

\section{Related Work}
Video summarization, as defined by Apostolidis et al. \cite{dnn-vs-review}, is the task of retrieving a ``synopsis" of a video by selecting the fewest and most pertinent parts of the video. Recent work favours the summaries creation in the form of video skims, due to user preferences \cite{dnn-vs-review}. The primary datasets which serve as benchmarks for the state-of-the-art are: TVSum \cite{tvsum} and SumMe \cite{summe}. Chhara et al. \cite{chhabra2023towards} is among the first works that introduces and evaluates the concept of ``fairness" within video summarization. In this context fairness addresses equal representations of individuals and protected groups within a final summary. In our work, we evaluate whether spatio-temporal relationships play a role or if these benchmarks may be statically biased. 
\subsection{Supervised Video Summarization with spatio-temporal models}
Among the earliest examples which utilised neural networks and spatio-temporal modeling with supervised learning was the approach introduced by Zhang et al. \cite{zhang2016video}. They employed the Long Short Term Memory (LSTM) model and formulated strategies to create optimization objectives from the provided annotations in the TVSum and SumMe datasets. Subsequent works built on top of it, while addressing the challenges posed by LSTMs in terms of modelling long range dependencies. These include approaches with Fully Convolutional Sequence Networks \cite{rochan2018video}, Memory networks \cite{memory-networks}, Graph networks \cite{sumgraph} and the predominant approach, Self Attention \cite{pglsum,NEURIPS2021_7503cfac,fajtl2018summarizing,MAAM,glrpe}.  
\subsection{Temporal Dependencies in Video Data}
Several works have reviewed whether deep learning models applied to video data truly learn spatio-temporal relationships. An example of this from Li et al. \cite{Li2018RESOUNDTA} in activity recognition, demonstrated that due to ``representation biases" within the dataset, models can ignore the temporal information within activity recognition benchmark datasets, UCF101 \cite{Soomro2012UCF101AD} and Kinetics \cite{kinetics} and rely only on static cues to classify activities. Li et al. \cite{li2019repair} also demonstrated a means to remove representation biases through dataset resampling.
Some works within activity recognition also manipulate the temporal order of frames/shots. Sevilla-Lara et al. \cite{sevilla2021only} highlighted a shuffling approach which aims to identify which videos require temporal information, dubbing them to be ``temporal classes". Huang et al. \cite{Huang2016DenselyCC} introduced two frameworks to isolate and analyse temporal features within popular activity recognition models.
\section{Methodology}
Our followed methodology consists of adapting an existing pipeline, using state-of-the-art models, which are trained using a series of temporal perturbation approaches.  
\subsection{Video Summarization Pipeline}
\label{sec:videosumpipe}
The video summarization pipeline used for this work is adapted from Zhang et al. \cite{zhang2016video}. In the following subsections we describe the formulatad pre-processing optimization and evaluation approaches. 
\subsubsection{Pre-processing}
Given a video $V$ with a sequence of $M$ frames, we first sub-sample it to a lower frame-rate, typically to $2$ frames per second. This step will lead to $N$, frames, denoted by $ N_j, j \in [1,2\;,3,\;.\;.\;.\;N] $ frames. Next, these frames are given to a feature extractor $F$. In this work we are using GoogleNet \cite{googlenet}, in order to enable a fair comparison with previous works. This results in a feature representation $F(N_j)$ per frame of the sub-sampled sequence, which are fed to the model for training/inference. 
\par
The TVSum and SumMe dataset annotations are pre-processed to create an optimization objective or ``ground truth" importance scores. These scores represent what a collection of human annotators deem to be relevant for a final summary. Both datasets provide multiple human annotations per video. For TVSum, this takes the form of a score between $1$ to $5$, while for SumMe, it is a score selected between $0$ and $1$. The ground truth importance score is then computed as the average over all annotators as mentioned by Fajtl et al. \cite{fajtl2018summarizing} in case of the TVSum dataset, while for SumMe, the scores are provided by the dataset creators.

\subsubsection{Model}
The whole or partial sequence of processed frames are given to the model to predict their ``importance scores". This results in a sequence of scores $\hat{y_j} = \text{Summariser}(F(N_j))$ where $\hat{y_j} \in (0,1)$, while the Summariser is a Deep Neural Network. This work provides the model with the full video as an input in line with previous work with self attention models \cite{li2022video,pglsum,fajtl2018summarizing}. The loss function is computed with respect to the ground truth importance scores per frame $y_j$ and depicted in equation \ref{eq:loss}:
\begin{equation}
\label{eq:loss}
    \mathcal{L} = \frac{1}{N}\sum_{j=1}^{N}(y_j - \hat{y}_j)^2
\end{equation}

\subsubsection{Evaluation metrics}
\label{sec:eval}
The models are evaluated by predicting the importance scores of all frames from a single video. The primary evaluation metrics used in video summarization are the Kendall and Spearman correlation coefficients as proposed by Otani et al. \cite{otani2018vsumeval}. This choice is due to the fact that the F1 score metric has been demonstrated to be greatly affected by the pre-processing and post-processing pipeline.
\par 
Due to the differences in the dataset annotations, the correlation coefficient is computed in two separate ways. In the case of the TVSum dataset, the correlation score of a video is computed as the average correlation over the model predictions with each human scaled score, since the annotations are scores between $1$ to $5$.
In the case of the SumMe dataset, as it provides only a $0/1$ score of whether a shot is included or excluded in a summary, we first compute the average score over all annotators and then we measure the correlation with the model's prediction.
For our experiments, we report and compare our results using only the Kendall correlation coefficient.

\subsection{Models}
\subsubsection{Temporally Invariant Models}
\label{sec:invariantmodel}

For this study, the Multi-layer Perceptron (MLP) and VASNet \cite{fajtl2018summarizing} are the chosen baselines. The VASNet model was chosen as it lacks positional encoding, since the authors noted that the positional order may not be relevant for video summarization \cite{fajtl2018summarizing}. However, they provide the model with the whole video as an input, which could already provide a degree of temporal context. Therefore, we utilise this model for testing our hypothesis with respect to temporal modeling. We investigate whether the frame-wise temporal dependencies within a video were important to achieve their success or whether VASNet could rely on spatial features alone to achieve the same performance. The MLPs architecture is described in the supplementary material.
\subsubsection{Adapted Models from Literature}
We utilise and adapt two models from the literature for our temporal perturbation experiment: VASNet \cite{fajtl2018summarizing} and PGL-SUM \cite{pglsum}. 
\paragraph{VASNet}
The VASNet architecture \cite{fajtl2018summarizing} utilises a self attention module alongside a regressor network to predict frame-wise importance scores. Their approach lacks positional encoding which renders the model permutation invariant. For our temporal perturbation experiment, we introduce absolute positional encoding added directly to the processed frame features prior to the self attention module. 
\paragraph{PGL-SUM}
The PGL-SUM model \cite{pglsum} also utilises self attention and positional encoding. This approach uses both local and global attention, fuses the information obtained from both attention branches and incorporates positional encoding directly to the attention matrix. Therefore, this model is an ideal candidate for the temporal perturbation, as by design it should be sensitive to temporal changes. 

\subsection{Temporal Perturbations}
\label{sec:perturbations}
We define two timescales of information as each video is comprised of disjointed shots, ``local" information within a shot and ``global" which pertains to the order of the shots. Therefore, we study perturbations across both of these defined scales. In this manner we address the question whether video summarization models are affected significantly either by the short term disruptions or by changes to the overall video structure. We formally define five strategies to perturb the order of the video frames for our experiments, described below. We also illustrate the effect different shuffling strategies have on the original sequence in Appendix \ref{sec:levenshtein}.

\subsubsection{Shuffling Strategies}
Let us consider a dataset composed of $NV$ videos (e.g. $NV = 50$ for TVSum). Each video $V^j$, $j \in \{1,\ldots,NV\}$ is comprised of $N^j$ frames, represented as an ordered set $V^j = \{ F_1,\;F_2,\;F_3,\; .\;.\;,F_{N^j}\}$ where $F_i,\; i\in N^j$ is the feature representation corresponding to the video frame with index $i$. 
\paragraph{Flip:} First, the ordered set is flipped leading to:  $V^j_f = \{ F_{N^j}, \;F_{N^{j-1}}\;,F_{N^{j-2}}, \; .\;. \;,F_1\}$. 
\paragraph{Fixed Segment Shuffles:} Next, the ordered set is divided into $M$ fixed segments having a length of $N^j/M$. For simplicity, lets assume a subset of the original video into a number of $M=3$ segments with length $M_l=2$, as $V = \{F_1,F_2 \;| F_3,F_4 \;| F_5,F_6 \}$. A fixed segment shuffle will permute the video as $V_fs = \{F_3,F_4 \;| F_5,F_6 \;| F_1,F_2\}$
\paragraph{Shot Level Shuffles:} Since the TVSum and SumMe datasets provide shot boundaries, we also utilise shot level shuffles\footnote{Since we utilise a sub-sampled input video, but the shot boundaries are defined for videos in the original frame-rate, we decide which frame belongs to which shot based on the frame index and the sampling rate. This choice is further discussed in the supplementary material}. Let's assume that given a video $V^j$ of $N^s$ shots, the corresponding representation will be $V^j  = \{ S_1,\; S_2, \;S_3, ...\; , S_{N^s}\}$. Furthermore, each shot is composed of $SF$ frames, such that  $S_i = \{F_1,F_2,F_3...F_{SF}\}$, where $F_j$ represents the frame level feature representation at index $j$. 
We propose three strategies that manipulate the shot order at various scales while keeping intact the order within the shots:
\begin{enumerate}
    \item \textbf{Intra-shot shuffling:} The overall shot order of the video is retained, but the frames within each shot are shuffled. For simplicity, for a video $V$ of four shots with varying length $V = \{F_1 ,F_2 \; | F_3,F_4,F_5,F_6 \; |F_7,F_8,F_9| \; F_{10},F_{11}\}$, the shuffled video appears  as:\\ $V_{is} = \{F_2,F_1|\; F_5,F_4,F_3,F_6|\; F_9,F_7,F_8 |\; F_{10}, F_{11}\}$.
    \item \textbf{Neighbouring Shot Shuffling:} Neighbouring shots are shuffled between each other, leading for example to: $V_{ns} =\{ S_2,S_1,S_3, \; | S_6,S_4,S_5 \; |\; . \;.  \; |S_{N},S_{N-1}\}$
    \item \textbf{Any shot shuffling:} This strategy is similar to the approach proposed within the fixed segment shuffles, while the difference consists of the varying size of a shot. Each shot is shuffled to randomly appear in a different position in the video.
\end{enumerate}

\section{Experiments}

We conducted two experiments. In the first, we estimated a baseline with temporally invariant models. This allows us a first sense of how much order contributes to the performance on benchmark datasets. In the second, we disrupted the temporal order of videos in different ways. This allowed us to investigate the impact of different types of temporal disruptions on performance. 

\subsection{Experimental protocol}
\paragraph{Datasets.} The \textit{TVSum} and \textit{SumMe} datasets are used for our experiments since they are the benchmarks in question. We utilised a pre-processed version of each of these datasets as provided by Zhang et al \cite{zhang2016video}. 
 In this paper, we report the results using the canonical data setting as described by Zhang et al \cite{zhang2016video}. The description of each dataset is provided in Table \ref{tab:datasetDescriptions}.
  \begin{table}[h]
        \centering 
    \caption{Datasets descriptions, according to Apostolidis et al. \cite{dnn-vs-review}}
    \begin{tabular}{|c|c|c|m{4cm}|} \hline 
         Dataset& Duration&  Videos& Topics\\ \hline 
         TVSum&  3 - 10&  25& news, how-to’s, user-generated,
documentaries\\ \hline 
         SumMe&  1-6&  50&  holidays, events, sports\\ \hline 

    \end{tabular} 
       \label{tab:datasetDescriptions}
 \end{table}


\paragraph{Experimental Design.}
The experiments were conducted using with the procedure followed in previous studies \cite{fajtl2018summarizing,pglsum,SumGDA,he2023a2summ,NEURIPS2021_7503cfac}. 
Each experiment is conducted using a five-fold cross validation split, where $80\%$ of the videos are used in the training split and $20\%$ of the videos are used in the test split from each of the benchmark datasets. The best correlation scores are recorded for each split and the overall performance is computed as an average over all the splits. We utilise $3$ permutations of a five-fold cross-validation split. 

\paragraph{Implementation Details.}
\label{sec:impdetails}
The pre-processing is adopted from Zhang et al. \cite{zhang2016video}, in which each video is sub-sampled to $2$ frames per second and each frame undergoes feature extract using GoogleNet \cite{googlenet} to. The full configuration for each of the models and each of the experiments can be found in the supplementary material. 
All models were trained for 50 epochs, with a weight decay of $1e^{-5}$, using the mean squared error loss function and gradient clipping. The temporally invariant baseline models are trained with a batch size of $128$, with a learning rate of $5e^{-5}$, while the temporal perturbation models are trained with a similar learning rate, but with a batch-size of $1$. 

\subsection{Description of experiments}

\paragraph{Experiment 1: Temporally invariant Baselines.} 
\label{sec:invariantprocedure}

We first establish a baseline for supervised video summarization relying purely on spatial features by removing any temporal context. Previous approaches \cite{zhang2016video,rochan2018video,fajtl2018summarizing,MSVA} typically give a part of the same video, or the whole video to the model for training. In our approach, we sample a batch of frames and ground truth annotations from any video in the training set, wherein the \underline{selected frames} in the batch can be from \underline{different videos and from any time-step}. This approach removes any potential temporal context, focusing entirely on the frame content. The model is optimised using the frames' ground truth scores and is evaluated in the same manner as done by previous existing works in video summarization (i.e the model is evaluated by providing the whole test video as an input). 
\par 


To compare between approaches, we train the MLP and VASNet models described in Section \ref{sec:invariantmodel} and we train two temporally aware models, namely, VASNet with positional encoding and PGL-SUM \cite{pglsum}. These models are trained using the original procedure as described in Section \ref{sec:videosumpipe} 

\paragraph{Experiment 2: Temporal Disruptions.}
We investigate the effect of temporal disruptions on the performance of a video summarization model. To demonstrate this, we first establish the baseline performance of a model trained on and evaluated with unshuffled data. Then, for each perturbation strategy as proposed in Section \ref{sec:perturbations}, we train the model on shuffled data and evaluate their performance on the unshuffled test split. The models used for this study are the PGL-SUM \cite{pglsum} and VASNet \cite{fajtl2018summarizing} with incorporated positional encoding.

\section{Results}
We conducted two experiments. In the first, we estimated a baseline with temporally invariant models. In the second, we disrupted the temporal order of videos in different ways. 
\begin{table}[h]
    \centering
        \caption{ Kendall Correlation coefficients from the Temporally Invariance Experiment}

    \begin{tabular}{|c|c|c|}
    \hline
        Model& TvSum &SumMe \\
        \hline
        Human Baseline\cite{otani2018vsumeval}&0.177& - \\
        \hline
        \multicolumn{3}{|c|}{Temporally Invariant Baseline}\\
        \hline
        VASNet (-PC)& 0.180 & 0.054\\
          MLP & 0.171 & 0.065 \\
          \hline
          \multicolumn{3}{|c|}{Temporally Dependent Models}\\
          \hline
         VASNet (+PC)& 0.147 & 0.082\\
         PGLSum \cite{pglsum}& 0.174 &0.033 \\
        
         \hline
    \end{tabular}
    \label{tab:perfs}
\end{table}
\subsection{Temporally Invariant Baseline}
The first experiment establishes the Kendall correlation coefficients that the temporally invariant models achieve on the two benchmark datasets for video summarization. We compare their results to existing work within video summarization, the human baselines provided by Otani et al. \cite{otani2018vsumeval} and with the existing works trained on our procedure to allow comparison. The results, as seen in Tables \ref{tab:perfs} and \ref{tab:compared_sota}, show that the MLP baseline is comparable to that of the human baselines ($0.171$ vs $0.177$) on the TVSum dataset. The self-attention model trained with the temporally invariant paradigm described in Section \ref{sec:invariantprocedure} also achieved $90\%$ of the performance of the state-of-the-art model MAAM \cite{MAAM} on the TVSum dataset (e.g. $0.180$ vs $0.207$). It is also worth noting that the Self Attention model trained using our temporally invariant paradigm described in Section \ref{sec:invariantprocedure} ($-PC$) outperformed its temporally dependent counterpart ($+PC$) by $19\%$ on the TVSum dataset. This result is notable as the self attention model with the temporally invariant paradigm received frames from different videos. This behaviour can be due to two possibilities, either that the frame level features alone are effective for the TVSum dataset, or that the use of positional encoding directly added to the CNN features may harm the models' prediction capability. 
\begin{table*}[h]
 \caption{Comparison of the Baseline Kendall Correlation Coefficients with other state-of-the-art models}
    \label{tab:compared_sota}
     \centering
     \begin{tabular}{|c||c|cc|}
     \hline
     Model &Split type& TvSum & SumMe  \\
     \hline
     \multicolumn{4}{c}{Existing Work}\\ 
     \hline
         A2Summ\cite{he2023a2summ} &1 $\times$ 5 FCV  & 0.137 &0.108 \\
          MAAM\cite{MAAM}& 1 $\times$ 5 FCV &0.179 &- \\
          MAAM(VIT)&1 $\times$ 5 FCV & 0.207 &0.227 \\ 
          SSPVS\cite{li2023progressive}& 1 $\times$ 5 FCV &0.181 &0.192 \\ 
          VHJMT\cite{li2022video}& 1 $\times$ 5 FCV &0.097 & 0.106\\ 
          Clip-It\cite{NEURIPS2021_7503cfac}& 1 $\times$ 5 FRV &0.108 &- \\ 
          SumGraph\cite{graph-reconstruction}&1 $\times$ 5 FCV  &0.094 & - \\ 
          PGL-SUM\cite{pglsum}& 1 $\times$ 5 FRV &0.150 & -\\
          MSVA\cite{MSVA} &1 $\times$ 5 FRV &0.190 & 0.200 \\ 
          \hline
          \multicolumn{4}{c}{Baselines}\\
          \hline
          MLP&3 $\times$ 5 FCV&0.171  & 0.065\\ 
          VASNet(-PC)&3 $\times$ 5 FCV&0.180  & 0.054\\
          \hline
          
     \end{tabular}
 \end{table*}
\par
However, in the case of the SumMe dataset, the results showcase that the VASNet with positional encoding(+PC in the table) narrowly beat VASNet without positional encoding(-PC in the table). This indicates that positional information may play a minor role within this benchmark dataset. 
\par 
\begin{figure}[h]
\centering
	   \begin{subfigure}[h] {0.49\textwidth}
        \centering
		\includegraphics[width=\linewidth]{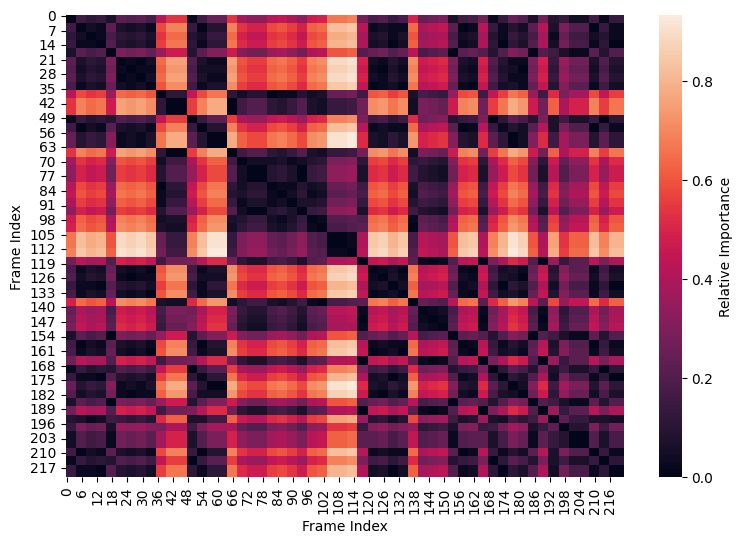}
  \caption{Ground Truth Differences}

		\label{fig:subfigrelimp1}
	   \end{subfigure}
    \hfill
	   \begin{subfigure}[h] {0.49\textwidth}
        \centering
		\includegraphics[width=\linewidth]{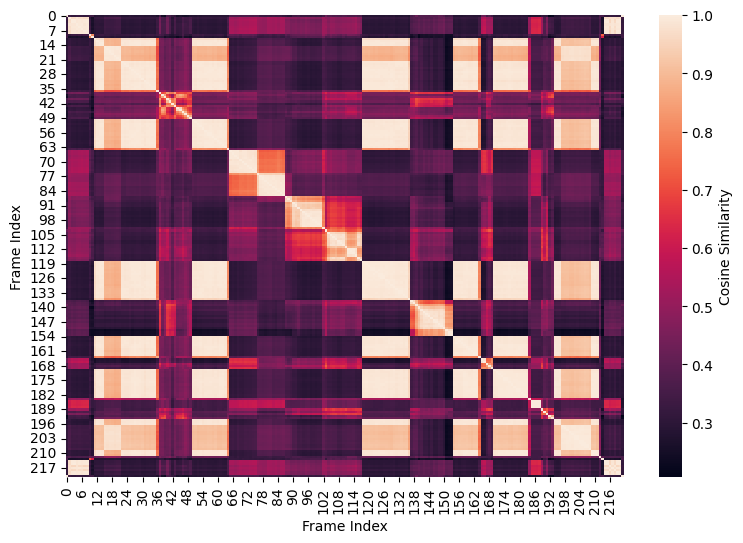}
  \caption{Framewise Cosine Similarity}

		\label{fig:subfigcossim1}
	    \end{subfigure}

     \caption{A visualization, for one video, of the heatmaps of the pairwise differences between the ground truth scores compared with the frame-wise cosine similarity. Here, frames that are visually similar also appear to have low differences in their importance (note, inverse color).}
    \label{fig:cosinesimilarity}
     \end{figure}

\par

To further investigate why these temporally invariant models achieve competitive Kendall coefficients, we analyse the processed datasets, in terms of features and labels to check whether there is any relationship between them. Our assumption is that similar features, measured in terms of their representation, will also be assigned similar importance scores. The exact followed procedure consists of creating two heatmaps, one including the pairwise cosine similarity between the features of each frame and the second encompassing the differences between their ground truth scores. We analyzed these relationships using with a few examples taken from the TVSum dataset and in Figure \ref{fig:cosinesimilarity}, ${Video}-5$ is depicted. It can be easily noticed that frames having a high cosine similarity also have small differences in their ground truth importance scores. In contrast, for $\text{Video}-32$,\footnote{Visualized in the supplementary material due to space limitations.} 
the rated importance difference did \textit{not} follow the frame-wise similarity.\footnote{This may be because $\text{Video}-32$ depicts a flash mob where each frame possesses similar objects and settings, but the actions and motions within the video are distinct between shots.} As $\text{Video}-5$ recorded a higher score than $\text{Video}-32$ for all of the models, this provides us with a possible explanation for the success of our temporally invariant models -- the nature of the datasets.  

\begin{table}[h]
    \centering
    \caption{The resulting Kendall Correlation Coefficients(best model in bold) of the temporal disruption experiment, for different models (PGL-SUM, VASNet+PC), and datasets (TVSum, SumMe).}
    \begin{tabular}{|c||cc|cc|}
    \hline
        Shuffle&\multicolumn{2}{c|}{PGLSum} &\multicolumn{2}{c|}{VASNet(+PC)} \\
        \hline
        & TvSum &SumMe &TVSum &SumMe\\
    Unshuffled &0.174 &0.033 &0.147 &0.082 \\
    Fixed Segment & \textbf{0.189} &0.085 &0.169 &\textbf{0.138}\\
    Flip & 0.176& 0.039&0.128 & 0.088\\
    IntraShot & 0.174&0.067&0.154&0.091\\
    Neighbour Shot & 0.175&0.067&0.164&0.128 \\
    Any Shot & 0.190&0.090& 0.171&0.125\\

 \hline
 \end{tabular}
  \label{tab:shuffleresult}
 \end{table}
\subsection{Temporal Disruptions}
    The second experiment demonstrates the effect that the temporal perturbations described in Section \ref{sec:perturbations} have on the performance of temporally dependent video summarization models. 
The results suggest that some of these strategies show little change over their unshuffled baselines, but show an improvement in strategies that shuffle across fixed time segments. As illustrated in Table \ref{tab:shuffleresult}, the intra-shot, flip, and neighbouring shot shuffles strategies score close to the TVSum baseline performance in both models. In the case of the SumMe dataset, they invariably improve the performance of the model. 
\par
The \textit{Fixed Segment Shuffle} and \textit{Global level shot shuffling} improve model performances across TVSum and SumMe. In particular, PGL-SUM shows an improvement on the Kendall correlation coefficient of $9.7\%$ using the \textit{Fixed Segment Shuffle} and $8.6\%$ using the \textit{Global level shot shuffling}  on the TVSum dataset. A notable point is that the shuffling strategies improved the correlation score over the SumMe dataset significantly in the case of VASNet with positional encoding, especially when using the Fixed Segment Shuffling, but not as significantly in the case of PGL-SUM.  

\begin{figure}[h]
\centering
	   \begin{subfigure}[h] {0.4\textwidth}
        \centering
		\includegraphics[width=\linewidth]{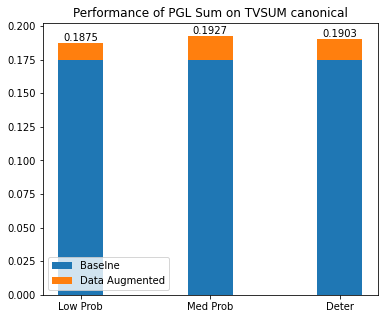}
  \caption{PGLSum: TVSum}

	   \end{subfigure}
    \hfill
	   \begin{subfigure}[h] {0.4\textwidth}
        \centering
		\includegraphics[width=\linewidth]{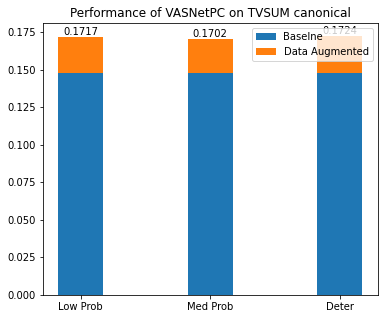}
  \caption{VASNet: TVSum}
\vfill
	    \end{subfigure}
    	   \begin{subfigure}[h] {0.4\textwidth}
        \centering
		\includegraphics[width=\linewidth]{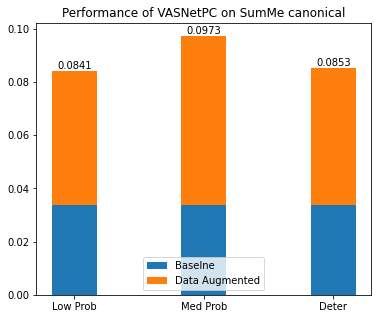}
  \caption{PGLSum: SumMe}

	   \end{subfigure}
    \hfill
	   \begin{subfigure}[h] {0.4\textwidth}
        \centering
		\includegraphics[width=\linewidth]{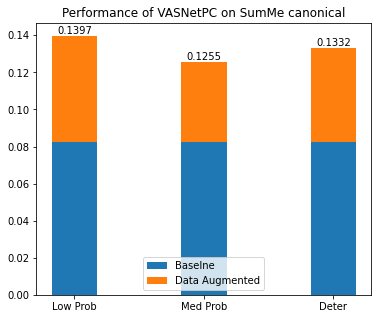}
  \caption{VASNet: SumMe}

	    \end{subfigure}

     \caption{The Results of the Data Augmentation Experiment. The orange shows the extent to which the models exhibited an improved Correlation Coefficient. As seen here, both PGLSUM and VASNet recorded an improvement, while this improvement was larger in the case of the SumMe dataset.}
    \label{fig:barchartresults}
     \end{figure}

\paragraph{Temporal Perturbation as Data augmentation.}
\label{sec:dataaug}

The improvements recorded on the Kendall correlation scores using the \textit{Fixed Segment Shuffles} and \textit{Shot Level Shuffles} strategies could be because they provide different views on the same data. One could assume that permuting fixed shots and segments provides multiple cuts over the same long video, presenting a different narrative each time. In other computer vision tasks such as image classification, such modifications like cutting parts of an image or flipping the image are used as data augmentation strategies to address data limitations. Along these lines, we tested whether these strategies can function as data augmentation since they appear to work similarly. As an extension to the previous experiment, we combine the flip and the fixed segment perturbations and present the model with a probability of receiving shuffled data. These two strategies were chosen as they introduce a change both globally (\textit{Flip}) and locally (\textit{Fixed Segment Shuffle}). The results in Figure \ref{fig:barchartresults} show that both models achieve an improved performance for both datasets. 
\par


    

\section{Discussion}

Given that the results without temporal information are already comparable to the human baseline ($0.170$ for the MLP model vs $0.177$ for the human baseline in terms of their Kendall correlation in the case of TVSum), we checked for which percentage of the predictions made by one baseline model (in this case the MLP) correlated well with human summaries. Assuming an acceptability threshold for the Kendall correlation informed by the human baseline of $0.15$, we see that this is achieved by $52\%$ videos of the TVSum dataset (26 out of 50) and $20\%$ of the SumMe dataset (5 out of 25). This indicates that a notable portion of each dataset may not require temporal context. However, this could also indicate that simply correlating between model predictions and human labels may not adequately measure a model's capacity to summarise videos based on context.

\par
The improvements seen when we introduce the \textit{Shot level Shuffles} and the \textit{Fixed Segment Shuffles} could indicate that short-term temporal context may benefit the model to a certain extent. This behaviour could be explained by the labeling strategy employed by the TVSum dataset, since the annotators were presented with video shots in a random order. The improvements recorded over the SumMe dataset especially highlight the data scarcity issue that plagues supervised video summarization, as a simple strategy improving the performance is quite indicative of this effect. 
\par

\section{Conclusion}

A key aspect of creating video summaries resides in deciding what should be included based on past and future context. Supervised video summarization models should learn to use temporal context to predict what is, and what is not, relevant to the final summary. The results indicate that temporal context provides a limited benefit towards supervised video summarization and that short temporal dependencies may be useful for the TVSum and SumMe benchmark datasets. More crucially, the results of our experiments suggest that models that lack temporal context achieve competitive scores on the video summarization benchmark datasets.  Jointly, our findings underscore the need for future work to concretely evaluate the potential static biases that may prevail in these benchmarks. More vitally, we need to consider the temporal bias not only when designing new methods, but also when we create new benchmark datasets. These new benchmark datasets for summarization should also consider temporal information.

\begin{acknowledgments}
  This publication is part of the project ROBUST: Trustworthy AI-based Systems for Sustainable Growth with project number KICH3.LTP.20.006, which is (partly) financed by the Dutch Research Council (NWO), RTL, and the Dutch Ministry of Economic Affairs and Climate Policy (EZK) under the program LTP KIC 2020-2023. 
\end{acknowledgments}


\appendix
\section{Differences between Shuffled and Original Video Sequences}
\label{sec:levenshtein}
We demonstrate the effect of different shuffles through the use of the Levenshtein/edit distance \cite{levenshtein1966binary}. The Levenshtein distance has been used previously to measure the similarity between two string sequences by quantifying the substitutions, insertions, and deletions to transform one sequence to another. Since our proposed shuffling strategy manipulates the order of a video sequence through the indices, the Levenshtein distance can be used to measure the dissimilarity between them. To demonstrate this aspect, we compute the Levenshtein distance over the entire TVSum and SumMe datasets for different shuffling strategies described in Section \ref{sec:perturbations}.
\begin{table}[h]
    \centering
    \caption{The Levenshtein distance for different shuffling strategies for the TVSum Dataset. A lower score implies that there has been a significant change between the sequences.}
    \begin{tabular}{|c|c|c|}
        \hline

         Shuffle type & Shuffle Iterations& Levenshtein Distance (scaled to 100)  \\
         \hline
         Flip &1 &  0.15 \\
         
         Intra Shot Shuffle  &3 &100.00  \\
         Fixed Segment Shuffle& 3& 11.03 \\
         Neighbouring Shot Shuffle& 3 &58.15\\
         Whole Shot Shuffle&  3&6.43 \\
         \hline
         \end{tabular}
    
    \label{tab:levdisttvsum}
\end{table}
\begin{table}[h]
    \centering
     \caption{ The Levenshtein distance of multiple shuffles for the SumMe Dataset. A lower score implies that there has been a significant change between the sequences.}
    \begin{tabular}{|c|c|c|}
        \hline

         Shuffle type & Shuffle Iterations& Levenshtein Distance (scaled to 100)  \\
         \hline
         Flip & 1&  0.19 \\
         
         Intra Shot Shuffle  &3 &100.00  \\
         Fixed Segment Shuffle&3 & 18.79 \\
         Neighbouring Shot Shuffle&3  &59.71\\
         Whole Shot Shuffle& 3 &13.22 \\
         \hline
         \end{tabular}
   
    \label{tab:levdistsumme}
\end{table}
It is important to note that the Levenshtein similarity is always 0 when comparing a flipped sequence with the original, and will also be 100 when comparing the intrashot shuffled sequence with the original. This is largely due to the way the Levenshtein distance is computed. However, it may not always be the case that the flipped video is semantically dissimilar from the original and that videos with their shots shuffled internally have no difference with their unshuffled counterparts. 
\section{Shot Division in Downsampled inputs}
Both the TVSum and SumMe datasets provided shot-boundaries which were created utilising the Kernel temporal segmentation algorithm. These shot boundaries were computed using the full sequence of inputs and given as indices where a shot starts or ends . For e.g Shot 1 could be between $(0,12)$ can be between the frame indices of $0$ to $12$. However, the processed sequences used for training and evaluation of the models were sub-sampled by skipping every fifteenth frame. Therefore, to assign which index belongs to which shot, we simply multiply each index by $15$ in the sub-sampled sequence and then assigned to which shot boundary in the original sequence based on the boundaries provided. An example of this is illustrated of this as follows; Assume we have a subsampled sequence of $9$ frames, lets arrange this as an index $[\;0,\;1,\;2,\;3,\;4,\;5,\;6,\;7,\;8\;]$, lets assume the shot boundaries provided by the original videos to be $(0,12),(13,60),(61,106),(107,120)$. Then we multiply the subsampled index by 15 $[0,\;15,\;30,\;45,\;60,\;75,\;90,\;105,\;120\;]$, then the final shots assigned are as $[0,\;1,\;1,\;1,\;1,\;2,\;2,\;2,\;2,\;3\;]$

\section{Experimental Configurations}
\subsection{Shared Hyperparameters}
All of the models were trained with the following set of hyperparameters shared between them
\begin{enumerate}
    \item Epochs: 50
    \item Weight decay: 1e-5
    \item Gradient Norm Clipping: 3 
    \item Learning rate: 5e-5
    \item Optimizer: Adam
    \item Input Dimensions: 1024
\end{enumerate}
\subsection{MLP and Attention}
For the temporally invariant baselines, the hyper-parameters for each model are listed as follows. The main observations relate to the batch size of 128 with a learning rate of 5e-5 and the ADAM optimizer to train the model.
\begin{itemize}
\item Self Attention configuration\begin{itemize}
     \item Self Attention Linear projection dimension: 1024
    \item FeedForward Neural Network dimensions: 1024
    \item Number of heads: 1
    \item Drop-out: 0.5
\end{itemize}
\end{itemize}
\subsection{PGL-SUM}
The PGL-SUM architecture was directly taken from the code provided by Apostolidis et al\cite{pglsum}. We take the configuration specified in the code which is as follows
\begin{itemize}
    \item Number of heads: \begin{itemize}
        \item Local attention: 4
        \item Global attention: 8
    \end{itemize}
    \item Number of Segments: 4
    \item Absolute positional encoding frequency : 10000
    \item Fusion Strategy: Addition
    \item Drop-out: 0.5
\end{itemize}
\section{Cosine Similarity versus Ground Truth Differences visualizations}
These are some of the visualizations of the frame-wise cosine similarity of the CNN features of a video versus the absolute differences between the ground truth importance score given in each of the datasets. We chose these examples to showcase the relationships between them 
\begin{figure}[h]
\centering
	   \begin{subfigure}[h] {0.49\textwidth}
        \centering
		\includegraphics[width=\linewidth]{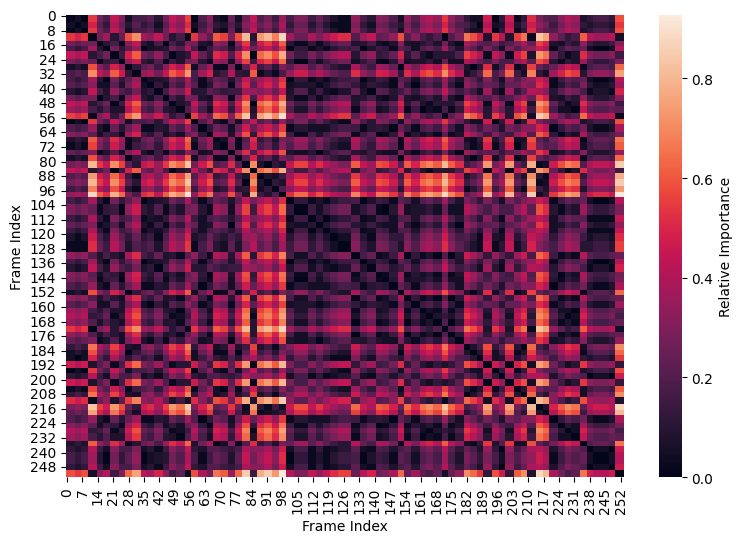}
  \caption{Ground Truth Differences}

		\label{fig:subfigrelimp2}
	   \end{subfigure}
    \hfill
	   \begin{subfigure}[h] {0.49\textwidth}
        \centering
		\includegraphics[width=\linewidth]{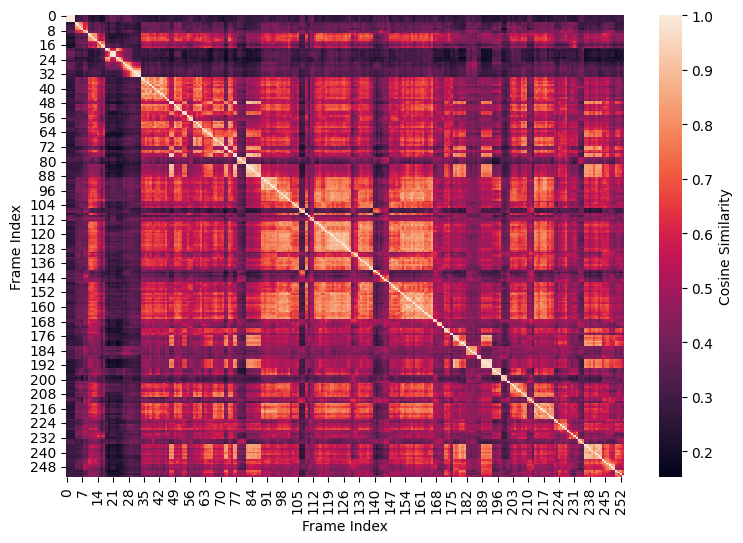}
  \caption{Framewise Cosine Similarity}

		\label{fig:subfigcossim2}
	    \end{subfigure}

     \caption{Heatmap of Video 32 of the TVSum Dataset}
    \label{fig:cosinesimilarity2}
     \end{figure}

\end{document}